\documentclass[10pt,conference]{IEEEtran}
\IEEEoverridecommandlockouts
\usepackage{cite}
\usepackage{amsmath,amssymb,amsfonts}
\usepackage{algorithmic}
\usepackage{graphicx}
\usepackage{textcomp}
\usepackage{xcolor}
\usepackage{subfig}
\usepackage{booktabs}
\usepackage[super]{nth}
\usepackage[normalem]{ulem}
\useunder{\uline}{\ul}{}
\usepackage{siunitx}
\usepackage{hyperref}
\def\BibTeX{{\rm B\kern-.05em{\sc i\kern-.025em b}\kern-.08em
    T\kern-.1667em\lower.7ex\hbox{E}\kern-.125emX}}
\begin{document}

\title{Extended Self-Critical Pipeline for Transforming Videos to Text (VTT Task 2021) -- Team: MMCUniAugsburg
}

\author{\IEEEauthorblockN{Philipp Harzig}
\IEEEauthorblockA{\textit{Multimedia Computing and Computer Vision Lab} \\
\textit{University of Augsburg}\\
Augsburg, Germany \\
philipp.harzig@uni-a.de}
\and
\IEEEauthorblockN{Moritz Einfalt}
\IEEEauthorblockA{\textit{Multimedia Computing and Computer Vision Lab} \\
\textit{University of Augsburg}\\
Augsburg, Germany \\
moritz.einfalt@uni-a.de}
\and
\IEEEauthorblockN{ Katja Ludwig}
\IEEEauthorblockA{\textit{Multimedia Computing and Computer Vision Lab} \\
\textit{University of Augsburg}\\
Augsburg, Germany \\
katja.ludwig@uni-a.de}
\and
\IEEEauthorblockN{Rainer Lienhart}
\IEEEauthorblockA{\textit{Multimedia Computing and Computer Vision Lab} \\
\textit{University of Augsburg}\\
Augsburg, Germany \\
rainer.lienhart@uni-a.de}
}
\maketitle

\begin{abstract}
The Multimedia and computer Vision Lab of the University of Augsburg participated in the VTT task only. We use the VATEX~\cite{wang2019vatex} and TRECVID-VTT~\cite{awad2019trecvid} datasets for training our VTT models. We base our model on the Transformer~\cite{vaswani2017attention} approach for both of our submitted runs, i.e., for run \textit{2021-01} . For our second model (\textit{2021-02}), we adapt the X-Linear Attention Networks for Image Captioning~\cite{pan2020x} which does not yield the desired bump in scores. For both models, we train on the complete VATEX dataset and $90\%$ of the TRECVID-VTT dataset for pretraining while using the remaining $10\%$ for validation.  \\
We finetune both models with self-critical sequence training~\cite{rennie2017self}, which boosts the validation performance significantly. Overall, we find that training a Video-to-Text system on traditional Image Captioning pipelines~\cite{vinyals2015show} delivers very poor performance. When switching to a Transformer-based architecture our results greatly improve and the generated captions match better with the corresponding video (see Figure~\ref{fig:example}). 
\end{abstract}


\section{Introduction}
In this notebook paper, we present our Video-to-Text model, which allows to create descriptions for arbitrary videos. Our model is inspired by the classical Transformer~\cite{vaswani2017attention} approach.

\section{Model}
\begin{figure}[ht]
	\centering
	\includegraphics[width=\columnwidth]{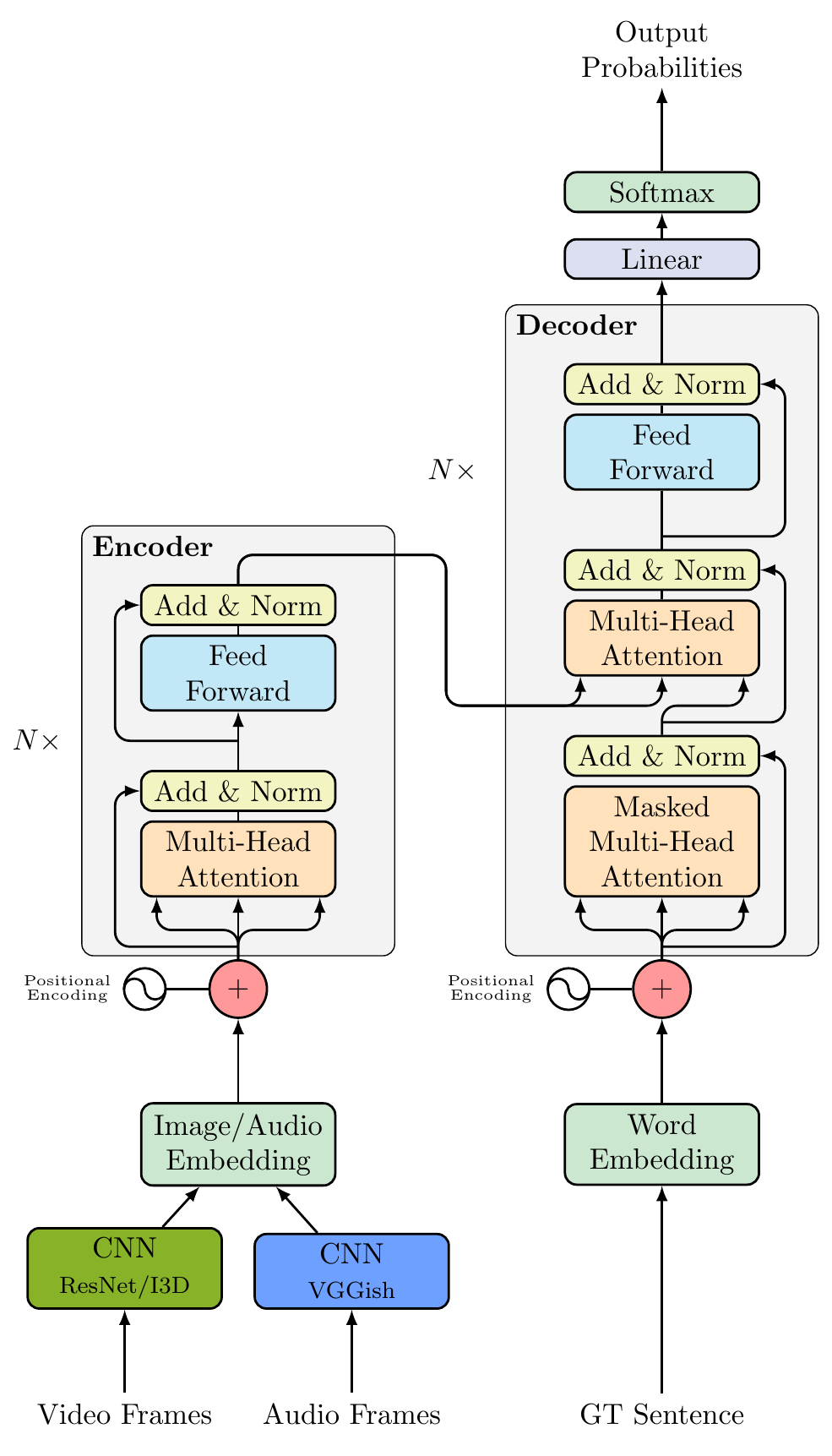}
	\caption{Our model architecture was slightly modified from the original Transformer~\cite{vaswani2017attention} to allow vision and audio frames as input to the encoder blocks. Model image inspired by \cite{vaswani2017attention} and modified to match our architecture.}
	\label{fig:model}
\end{figure}

\subsection{Preprocessing of Videos}

\textbf{Single Images.}
In order to process the videos in our model, we first need to extract single frames. We use ffmpeg for extracting every frame of each video of the respective dataset. We use ResNet-101~\cite{he2016identity} to compute features for the extracted frames. More specifically, we resize the input images to $224\times 224$ and use the average pooled features with dimension $\mathbb{R}^{2048}$.

\vspace{\baselineskip}
\textbf{I3D Features.} We additionally extract features with the Inflated 3D ConvNet (I3D)~\cite{carreira2017quo} similar to frame-level features. Instead of forwarding frame images through the ResNet-101~V2 network, we extract video clip features with the RGB-I3D pretrained on the Kinetics Human Action Video dataset~\cite{kay2017kinetics}. 

\vspace{\baselineskip}
\textbf{Audio features.} We take the audio of the video, resample it to \SI{16}{\kilo\hertz} and extract features with the VGGish~\cite{hershey2017cnn} network. If no audio stream for a video is existent, we create a dummy feature vector with all zeros.

\subsection{Preprocessing of Tokens}
In contrast to our 2020 submission, we do not employ a default tokenizer, but we use the WordPiece Tokenizer~\cite{wu2016google} to generate the tokens. We load pretrained embedding weights\footnote{\url{https://tfhub.dev/google/small_bert/bert_uncased_L-8_H-512_A-8/1}} from the BERT\textsubscript{SMALL} model.

\subsection{Model}
An overview of our model architecture is depicted in Figure~\ref{fig:model}. In comparison with the original Transformer~\cite{vaswani2017attention} architecture, we changed the encoder part to accept image features instead of embedded words.
That is, we exchanged the sentence encoder with a video encoder. More specifically, we replaced the input embedding with an image embedding, which is standard practice in common image captioning models~\cite{vinyals2015show}. An image embedding layer embeds the image features into the desired embedding space. In our model, we use ResNet-101 features $\in \mathbb{R}^{2048}$ and embed them into the encoder space with dimension $d_{\text{model}}=512$. Additionally, we concatenate audio features extracted by the VGGish~\cite{hershey2017cnn} network. We use a seperate embedding layer for the audio features.\\
We also use positional encoding to encode the order of every single frames in the video. As the Transformer architecture does not care about the order of the input, i.e., every frame can influence every other frame in the same way, we need to explicitly tell the encoder the frame number. Similar to the original paper, we use a positional encoding to encode the frame number, which we add on top of the embedded image features. 
The sequence length of image features or I3D features is varying. Therefore, we cannot simply concatenate vision and audio features as the added positional encoding may signal the encoder that it receives a vision feature as input when in reality it is an audio feature.
We assume a fixed starting position for all audio features which we set to $300$ (i.e., there are no more than $300$ image frames for any video in the dataset). Finally, we add positional encodings for indexes $[300, 301, \dots]$ on top of the embedded audio features.\\
In the encoder, we make use of the memory-augmented encoding~\cite{cornia2020meshed}, which encodes multi-level visual relationships with a priori knowledge. 
In the original work, Cornia et al. use a persistent, learnable memory vector which is concatenated to the key and value of the self-attention blocks of the Transformer. 
These memory vectors allow to encode persistent a-priori knowledge about relationships between image regions. In contrast to the original work, we work with video sequences instead of still images with regions. Adapted to our architecture, the memory vector encodes a-priori knowledge about relationships between frames in a given video. We did not change the architecture of the decoder block (see Section~\ref{tab:train-details}).

\section{Datasources}

We use two datasets for training our models, which are described below. Additionally, we show some dataset statistics in Table~\ref{tab:datasets}.
Note that we also trained on AC-GIF~\cite{pan2020auto} and MSR-VTT~\cite{xu2016msr} in last year's challenge. However, we found that VATEX delivers far better and more consistent results.

\begin{table}[tbp]
	\resizebox{\columnwidth}{!}{
	\begin{tabular}{@{}lllll@{}}
		\toprule
		Dataset & \# Videos (clips) & \# Sentences & \# Videos avail. & \# Sentences usable \\ \midrule
		VATEX~\cite{wang2019vatex} & \num{41269} & \num{349910} & \num{38109} & \num{323950}  \\
		MSR-VTT~\cite{xu2016msr} & \num{10000} & \num{200000} & \num{7773} & \num{155460} \\
		TRECVID-VTT~\cite{2021trecvidawad} & \num{7485} & \num{28183} & \num{5971} & \num{22547} \\
    	AC-GIF~\cite{pan2020auto} & \num{163183} & \num{164378} & \num{163183} & \num{164378} \\
		\bottomrule
	\end{tabular}
	}
	\caption{Different datasets and their respective number of video clips and number of available videos. Sentences are available for every video, however, not every video was available to be downloaded from YouTube.}
	\label{tab:datasets}
\end{table}

\subsection{TRECVID-VTT}
We use the official TRECVID-VTT dataset~\cite{2021trecvidawad} which contains videos from the TRECVID VTT from 2016-2019. We only use the Twitter Vine subset of videos. In total, this subset contains $6,475$ videos from which we use $5,971$ available videos with $22,547$ captions. In all our experiments we train on $90\%$ and validate the model on $10\%$ of the videos.

\subsection{VATEX}
We additionally train on the VATEX Dataset~\cite{wang2019vatex} to boost the performance of our final models.
We trained our last-year models on MSR-VTT~\cite{xu2016msr}, but found that the MSR-VTT dataset is less representative than the VATEX dataset and results in lower scores.
The VATEX dataset is split into 4 sets, i.e., the training set, the validation set, the public test set and the private test set. The VATEX dataset comes with 10 English and 10 Chinese captions per video clip. Most video clips have a length of \SI{10}{\second}. 

\begin{table*}[ht!]
	\centering
	\caption{Submitted models (in bold) and their respective validation scores. We validated all of our models after every epoch on $10\%$ of the TRECVID-VTT dataset to select a model to submit. We also include our models from last year (2020-01-ft and 2020-02-ft) for comparison.}
	\label{tab:val-results}
	\begin{tabular}{@{}llccccclll@{}}
		\toprule
		\multicolumn{1}{c}{\textbf{Model}} & \multicolumn{1}{c}{\textbf{epochs}} & \textbf{ft} & \multicolumn{1}{l}{\textbf{Features}} & \multicolumn{1}{l}{\textbf{$|\textrm{mv}|$}} & \multicolumn{1}{l}{\textbf{Vocabulary}} & \multicolumn{1}{l}{\textbf{lr Schedule}} & \multicolumn{1}{c}{{\textbf{B-4}}} & \multicolumn{1}{c}{\textbf{C}} & \multicolumn{1}{c}{\textbf{M}} \\ \midrule
		2020-01-ft & 25 & \checkmark & CNN & 64 & Default & Default & 0.076 & 0.176 & 0.116 \\
		2020-02-ft & 1 & \checkmark & CNN & 64 & Default & Default & 0.061 & 0.151 & 0.110 \\
		\midrule
		2021-01 & 43 & --- & I3D & 64 & WP-BERT & sgdr & 0.101 & 0.249 & 0.249 \\
		\textbf{2021-01-ft} & 3 & \checkmark & I3D & 64 & WP-BERT & $5\cdot10^{-6}$ & 0.142 & 0.308 & 0.160 \\
		2021-02 & 15 & --- & I3D & 64 & WP-BERT & sgdr & 0.109 & 0.226 & 0.137 \\
		\textbf{2021-02-ft} & 0.33 & \checkmark & I3D & 64 & WP-BERT & $5\cdot10^{-6}$ & 0.115 & 0.244 & 0.142 \\ \bottomrule
		\end{tabular}%
\end{table*}
\section{Model configurations}
We submitted two models for the Video-to-Text (VTT) task. Both of our models are pretrained on a merged dataset and then finetuned on the merged dataset as well. \\
For our primary model (cf. \textit{2021-01}), we first train a base model on the full MSR-VTT dataset and $90\%$ of the TRECVID-VTT dataset. We select the model by employing an early-stopping strategy on the CIDEr score of the remaining $10\%$ of the TRECVID-VTT dataset.
In contrast to last year's primary model (\textit{2020-01}), we train on I3D features instead of ResNet features. Furthermore, we add audio features for the VATEX part of our training set (the TRECVID dataset does not come with audio) and train the model with a modified learning rate schedule.
For finetuning, we use the base model and train it on the same dataset, but enable self-critical sequence learning~\cite{rennie2017self} with a constant learning rate $\eta=5\cdot10^{-6}$. 
Here, we calculate the CIDEr scores for a baseline caption $\hat{w}$ and sample 5 additional captions $w^s$, respectively. Subsequently, we can baseline the reward of the sampled captions by subtracting the CIDEr score for the baseline caption.
As a consequence, sampled captions with a higher CIDEr score than the baseline caption get a positive reward and vice versa.
The gradient of the loss function can be approximated as follows:
\begin{equation}
    \nabla_{\theta}L(\theta) \approx -(r(w^s)-r(\hat{w})) \nabla_{\theta} \log_{p_{\theta}}(w^s).
\end{equation}
Each word will be weighted according to its $\log$ probability and $r(\cdot)$ is the reward function. $\theta$ are the parameters of the network and define a policy $p_{\theta}$.
For our final models, we additionally optimize the BLEU-4 metric. Therefore, our reward function becomes
\begin{equation}
    r(\cdot) = \lambda_{\textrm{CIDEr}} \cdot r_{\textrm{CIDEr}}(\cdot) + \lambda_{\textrm{BLEU-4}} \cdot r_{\textrm{BLEU-4}}(\cdot),
\end{equation}
where $\lambda_{\cdot}$ is a weight for the corresponding metric. \\
Our second model (cf. \textit{2021-02}) is trained similarly, except we implemented X-Linear Attention~\cite{pan2020x}. We fine-tune this model in the same way as model \textit{2021-01}. In Table~\ref{tab:train-details}, we present the number of training samples used for training the base and finetuned models.\\
Our models use 8 encoder and 8 decoder blocks. We use 8 attention heads and a model dimension of $d_{\text{model}}=512$. For the position-wise feed-forward networks, we set $d_{ff}=2048$ as the inner-layer dimensionality.
We use a memory-vector size of $d_{\text{memory}}=64$. The primary model use the default BERT subtoken vocabulary with $30522$ subword tokens. It does not use complete words for the vocabulary, but tries to build words from subwords, i.e., it splits words into subwords if a word is not in the initial dictionary.

\begin{table}[tbp]
	\caption{Data source used for training our models. We also depict the total number of training and validation samples used. }
	\label{tab:train-details}
	\resizebox{\columnwidth}{!}{
		\begin{tabular}{@{}lll@{}}
			\toprule
			Model: Data sources                           & \# train samples & \# val samples \\ \midrule
			1: MSR-VTT + $90\%$ VATEX           & \num{273314}                 & \num{3602}              \\
			\bottomrule
		\end{tabular}
	}
\end{table}
\section{Training}
\begin{figure}[h]
	\centering
	\includegraphics[width=\columnwidth]{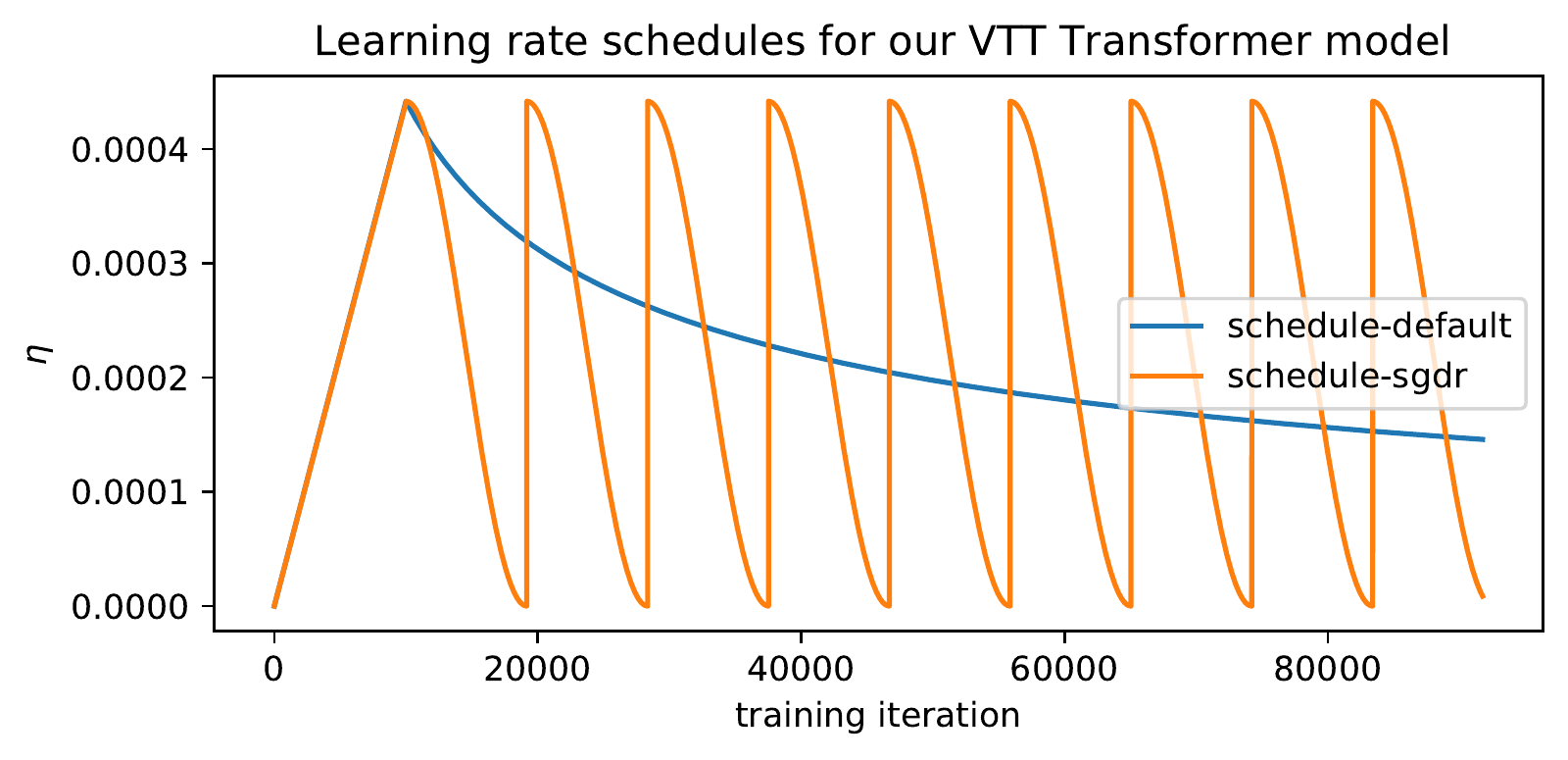}
	\caption{The blue line shows the default learning rate schedule for a Transformer with 10,000 warm-up steps. The orange line shows our learning rate schedule, a combination of SGDR~\cite{loshchilov2016sgdr} and the warm-up phase.}
	\label{fig:schedules}
\end{figure}

We train our models in a multi GPU setting, i.e., we train the model on 4 NVIDIA Tesla A100 GPUs simultaneously. We use a batch size of 128 per GPU, resulting in an effective batch size of 512. We use the Adam~\cite{kingma2014adam} optimizer with $\beta_1=0.9,\beta_2=0.98$ and $\epsilon=10^{-9}$. Similar to \cite{vaswani2017attention}, we train with a variable learning rate $\eta$ over the course of the training (\textit{schedule-default}). However, we combine the original learning rate with SGDR (Stochastic Gradient Descent with Warm Restarts, \textit{schedule-sgdr})~\cite{loshchilov2016sgdr} learning rate schedule. We plot the SGDR learning rate schedule combined with a warm-up phase in Figure~\ref{fig:schedules}. In contrast to the original Transformer architecture, we used $w = 10,000$ for the number of warm-up steps. \\
For the base model of our primary model (\textit{2021-01}), we observed the best validation performance on TRECVID-VTT after $43$ epochs with a CIDEr score of $0.249$. We used this model to finetune with self-critical sequence learning (\textit{2021-01-ft}). In doing so, we significantly improved the scores as can be seen in Table~\ref{tab:val-results}. For our second model (\textit{2021-02}), we chose the same approach but trained the base model with a transformer that employs X-Linear Attention~\cite{pan2020x}. The best scores were also observed after $15$ epochs and are in the same range as our primary model. However, when finetuning the second model with self-critical sequence learning, the scores did improve in contrast to the base model, but our primary model performs better. We submitted results generated by our models \textit{2021-01-ft} and \textit{2021-02-ft}, because we selected it based on the CIDEr scores and the generated captions on the validation set looked quite promising.



\section{Results}
\begin{figure*}[ht]
    \subfloat{
	       \includegraphics[width=\linewidth]{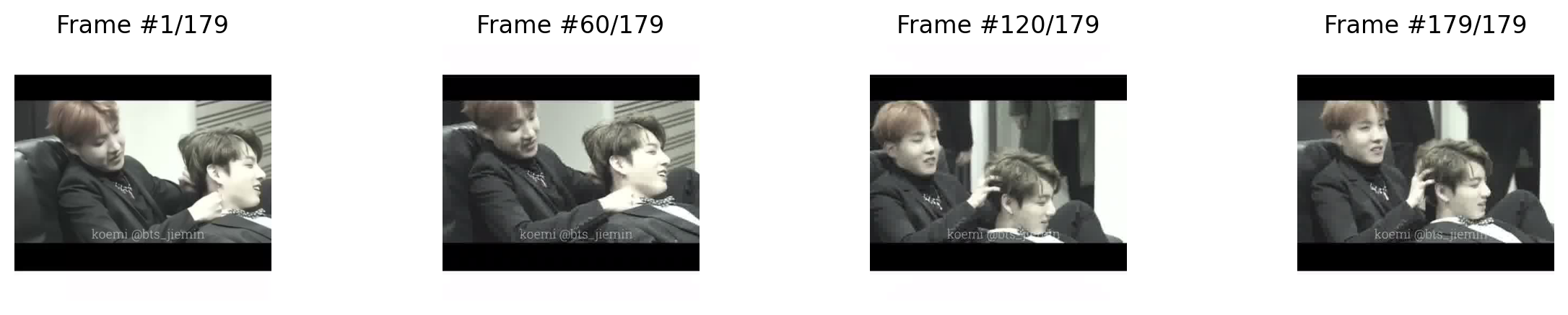}
	} \\
    \footnotesize{
        \input{sample_gens/trecvid2021_sample_video_TRECVID_VTT_2021_05042_caps.txt}
    }
    \subfloat{
	       \includegraphics[width=\linewidth]{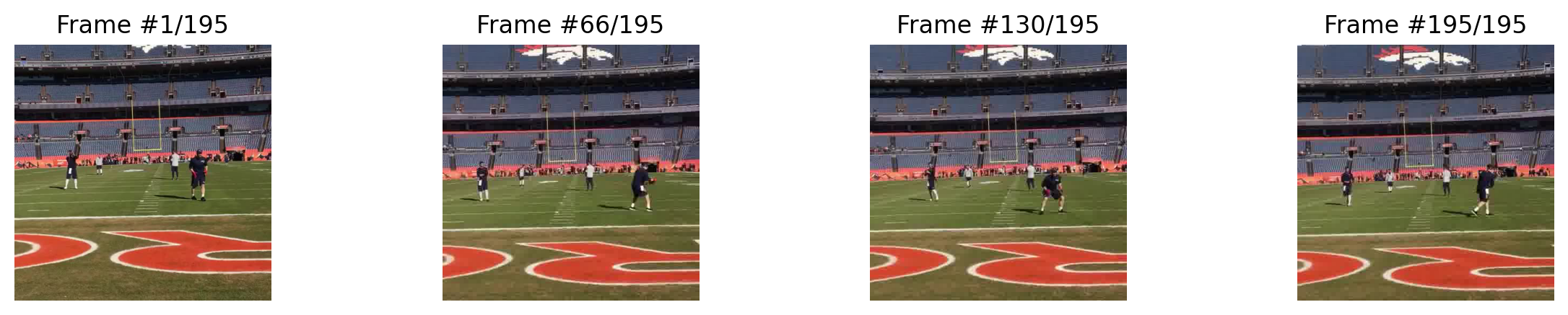}
	} \\
    \footnotesize{
        \input{sample_gens/trecvid2021_sample_video_TRECVID_VTT_2021_05875_caps.txt}
    }
    \subfloat{
	       \includegraphics[width=\linewidth]{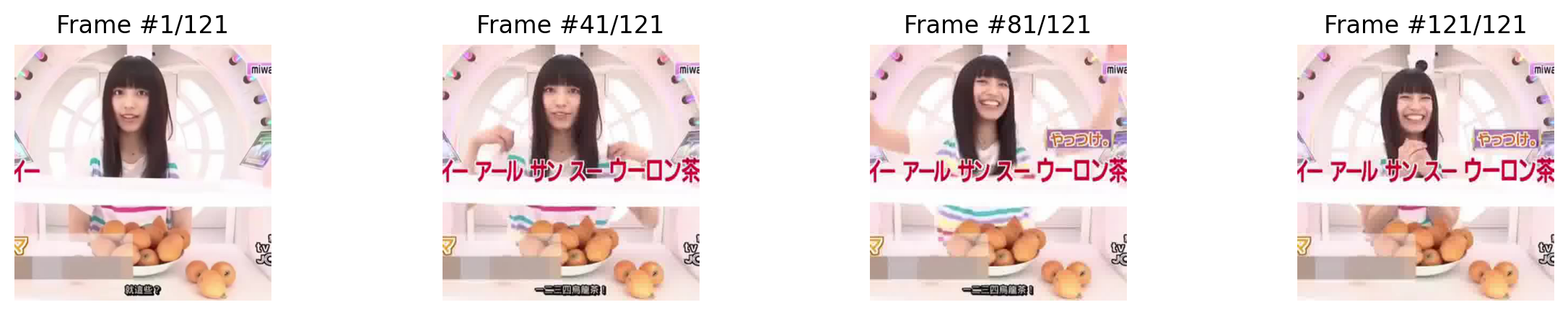}
	} \\
    \footnotesize{
        \input{sample_gens/trecvid2021_sample_video_TRECVID_VTT_2021_04267_caps.txt}
    }
    \subfloat{
	       \includegraphics[width=\linewidth]{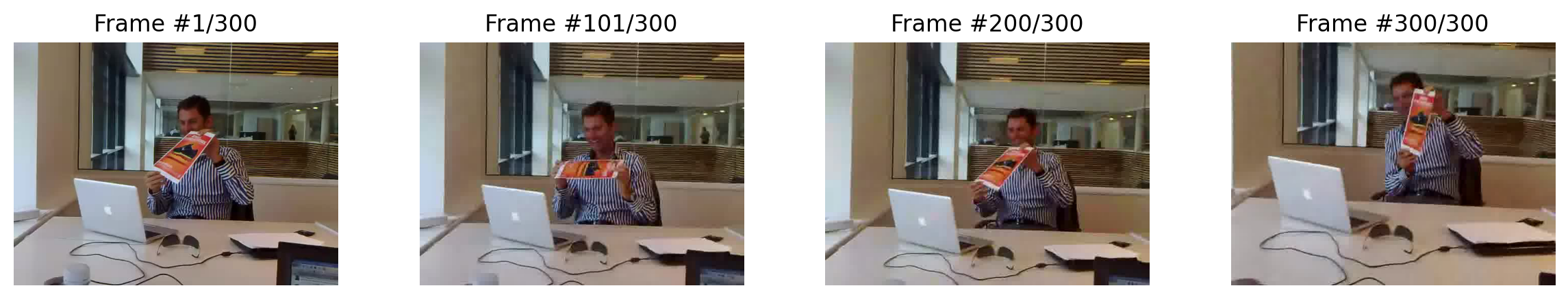}
	} \\
    \footnotesize{
        \input{sample_gens/trecvid2021_sample_video_TRECVID_VTT_2021_06503_caps.txt}
    }
    \subfloat{
	       \includegraphics[width=\linewidth]{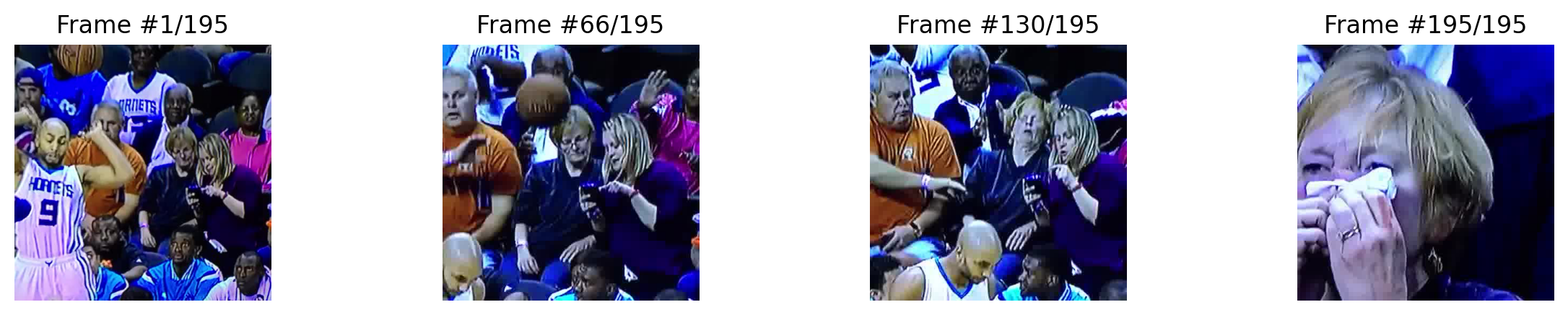}
	} \\
    \footnotesize{
        \input{sample_gens/trecvid2021_sample_video_TRECVID_VTT_2021_00208_caps.txt}
    }
	\caption{Five videos from the validation dataset and the corresponding captions generated by our models.}
	\label{fig:example}
\end{figure*}

\begin{table}[ht!]
	\caption{Submitted models and their respective performance on the unseen test dataset. Models with (1977) denote performance on the extended unreleased test set from 2021. The other models are comparable to last year's results (seen in the first two rows).}
	\label{tab:final-results}
	\resizebox{\columnwidth}{!}{
		\begin{tabular}{@{}lllll@{}}
			\toprule
			\textbf{Model} & \textbf{BLEU} & \textbf{CIDEr} & \textbf{CIDEr-D} & \textbf{METEOR} \\ \midrule
			\textbf{2020-01-ft} & 0.018 & 0.140 & 0.064 & 0.202 \\
			\textbf{2020-02-ft} & 0.011 & 0.136 & 0.060 & 0.204 \\
			\midrule
			\textbf{2021-01-ft} & 0.022 & 0.315 & 0.180 & 0.292 \\
			\textbf{2021-02-ft} & 0.015 & 0.247 & 0.137 & 0.260 \\
			\textbf{2021-01-ft (1977)} & 0.022 & 0.313 & 0.178 & 0.292 \\
			\textbf{2021-02-ft (1977)} & 0.015 & 0.246 & 0.137 & 0.260 \\ \bottomrule
			\end{tabular}%
	}
\end{table}
For the TRECVID 2020 workshop~\cite{2020trecvidawad}, we submitted captions generated on the provided test videos ($1,700$) for basic transformer models (see last year's notebook paper for details~\cite{DBLP:conf/trecvid/HarzigELL20}). In a nutshell, we implemented a vanilla Transformer that accepts only image features from a ResNet with support for memory-augmented vectors~\cite{cornia2020meshed}. \\
For this year's workshop~\cite{2021trecvidawad}, we extended our model to support audio frames, features from the Inflated 3D ConvNet (I3D)~\cite{carreira2017quo} and self-critical sequence training. We submitted captions generated by our two finetunes models (2021-01-ft and 2021-02-ft). \\
These captions were evaluated by the workshops organizers. Compared to our validation set scores, the evaluation on the test set yields worse results as can be been in Table~\ref{tab:final-results}. Especially, the BLEU score is much lower on the test data than on the evaluation data. \\
We depict five videos and their generated caption in Figure~\ref{fig:example}. We see that for the first three videos our generated captions from the model \textit{2021-01-ft} match the video content quite good. 
The first video description is correct.  Only if we look closer, we see that one person is giving the other person a massage.
In the second video, our model detects correctly that we see a football field and a group of people which are indeed playing footbal.
In the third video, our model detects a young woman who looks into a camera. However, it fails to detect that the woman is cheering in back of some apples. For the fourth video the model correctly detects a man. But the man is not reading a book, rather he is showing an ad in front of his notebook. In the fifth video, the model detects that there is basketball game going on. However, it shows the audience rather than basketball players sitting on a bench. But in the first frame, we see a basketball play, hence, the model may take this as a hint for generating the sentence.

\section{Conclusion}

In this notebook paper, we presented our VTT model based on a Tansformer~\cite{vaswani2017attention} architecture. By extracting features for every frame of the videos, we were able to adapt the Transformer architecture to use videos in the encoder block. Furthermore, we extracted features with the I3D network that is may extract contextual information related to the time-axis of the video.
In addition, we modified the Multi-Head Attention of the encoder to use memory vectors similar to \cite{cornia2020meshed} which allow to memorize a priori knowledge about relationships between video frames. Finally, we finetune our models with self-critical sequence learning that directly optimizes the CIDEr and BLEU-4 metrics. Thus, we generate captions that describe video contents (see Figure~\ref{fig:example}). However, as not all objects and circumstances of the videos are detected and described correctly, we want to address object and relationship detection in future work.

\bibliographystyle{ieeetr}
\bibliography{trecvid2021} 

\end{document}